# Prediction of Auto Insurance Risk Based on t-SNE Dimensionality Reduction

**Joseph Levitas**  j.levitas@kasko2go.com

**Konstantin Yavilberg**  k.yavilberg@kasko2go.com

**Oleg Korol**  o.korol@kasko2go.com

**Genadi Man**  g.man@kasko2go.com

## Abstract

**Correct risk estimation of policyholders is of great significance to auto insurance companies. While the current tools used in this field have been proven in practice to be quite efficient and beneficial, we argue that there is still a lot of room for development and improvement in the auto insurance risk estimation process. To this end, we develop a framework based on a combination of a neural network together with a dimensionality reduction technique t-SNE (t-distributed stochastic neighbour embedding). This enables us to visually represent the complex structure of the risk as a two-dimensional surface, while still preserving the properties of the local region in the features space. The obtained results, which are based on real insurance data, reveal a clear contrast between the high and the low risk policy holders, and indeed improve upon the actual risk estimation performed by the insurer. Due to the visual accessibility of the portfolio in this approach, we argue that this framework could be advantageous to the auto insurer, both as a main risk prediction tool and as an additional validation stage in other approaches.**

*Keywords*--- Auto insurance, Machine learning, Neural network, Risk estimation, t-SNE

## 1. Introduction

One of the main challenges of auto insurance companies is the prediction of a policyholder's risk with the goal of optimizing its premium and reducing the insurance claims. Ideally, at the time of forming a policy, the insurer would need to know the likelihood of receiving a claim and its severity. These, of course, cannot be accurately predicted as there are numerous factors contributing to the result, many of which are unknown to the insurer. Traditionally the main tools of the insurer are based on the method of *Generalized Linear Models* (GLM's) [1], [2], [3]. The typical approach is fairly straightforward: it is assumed that the data follows one of several probabilistic models, usually a distribution in the exponential family, each associated with a certain pricing mechanism. Despite the popularity and simplicity of the GLM's, in the last several years the field of auto insurance has seen a movement towards more advanced machine learning methods. These methods significantly increase the class of functions that can be approximated compared to the GLM's as they are not restricted to a specific probabilistic distribution. They are more efficient in handling large and complex data sets, as well as fitting non-linearities to the data.

Several authors have used different approaches in order to address the issue of claim and risk prediction. Guelman modelled and predicted the loss-cost via a tree-based gradient boosting method [4]. Yeo et al. employed both regression and hierarchical clustering in order to obtain the claim cost in each risk group [5]. Quan & Valdez predicted the claim using multivariate decision trees [6]. Although much of the research focuses on the claim and risk, some researchers have looked at different insurance metrics. Chapados et al. proposed a pricing strategy by using a mixture of several neural networks to estimate the insurance premiums [7]. Smith et al. used various regression and classification models to analyse customer retention patterns [8]. Staudt & Wagner used random forests to predict the claim severity in cases of car collisions [9]. Although many of these studies have achieved promising results, the issue of building an interpretable and visually accessible model still remains. This objective is particularly important in a business environment, where models must generally be approved by decision makers who are inexperienced in statistical analysis but nevertheless need to understand how the risk is being estimated.

To address the challenge of visually interpretable results, we explore the combined framework of a neural network (NN) with a dimensional reduction technique. The method of t-distributed stochastic neighbour embedding (t-SNE), developed by van der Maaten & Hinton [10], can efficiently project complex data sets onto a two-dimensional plane, while preserving as much as possible the local structure present in the original high-dimensional space. The use of t-SNE in supervised learning is a fairly novel concept, and it was studied and implemented only in recent years. Both regression and classification settings have been studied, where the t-SNE was used in combination with algorithms such as k-NN, CNN and random forests [11], [12], [13], [14], [15], [16], [17], [18]. These hybrid models have been shown to give promising results with performance comparable to methods which do not implement dimensional reduction and thus lack the crucial aspect of interpretability.

The method we propose here employs a three-stage process to obtain a two-dimensional risk surface. First, we implement t-SNE to obtain a dimension-reduced features space, which is then trained as a target via a neural network. Finally, an additional neural network is used to estimate the risk for each policy where the features space is the t-SNE 2D plane. Considering the stochastic nature of the t-SNE, one would assume that using it in a supervised learning setting is not optimal. However, due to the structure of the risk surface and the class-imbalanced nature of insurance claims, the results we obtain are quite robust. Beyond the predictive capabilities of our approach, it also serves as a powerful visualization and validation mechanism. Since our method generates a two-dimensional risk surface, one can exploit it to achieve higher confidence in the validation of any general algorithm for risk prediction by a simple visual examination.

The objective of this paper is to outline the method and show its application to the analysis of auto insurance risk. The remainder of the paper is organized as follows: In section 2 we present the data and describe the model used throughout. In section 3 we implement the method on a test case and present the results, where a thorough analysis and application of the method are also given. Section 4 presents a final discussion of the results and future directions.

# 2. Model Description

## Risk Estimation Problem Definition

The task we address in this paper is the estimation of an auto insurance claim risk associated with a new candidate. For our case study we use a real-life data from a large insurance company. The dataset consists of 30000 insurance contracts, and 14 parameters (features) associated with each contract. These are:

1. Policyholder's home latitude and longitude (2 features)
2. Car price
3. Engine power
4. Policyholder's age and License age (2 features)
5. Vehicle age
6. Vehicle type (7 features)

The first 7 features (Nos. 1-5) are continuous variables, whereas the last 7 (No. 6) are a one-hot-encoded binary versions of the different vehicle types. In addition, we have a binary claim/no-claim indicator associated with each contract, which was used as a target (forecasting) variable. As is usual in the field of auto insurance, the claims distribution results in a class-imbalanced target variable. All features were normalized to obtain a zero mean and a unit variance. Thus, the standard task is essentially an $R^{14} \rightarrow [0,1]$ function estimation, where 0 and 1 correspond to absence and to appearance of a claim, respectively. This mapping results in a multi-dimensional function, which can be used to assess the risk of new policyholders. It can be carried out by a variety of well-known classification or regression methods [19]. Unfortunately, it's inner structure (in the $R^{14}$ space) usually cannot be observed or understood by the most curious user.

As stated in the introduction, this paper aims to achieve a solution which can be easily observed, and thus could be intuitively understood. This attribute of the solution we call *Visibility* and it must be distinguished from *Explainability*, which represents a different aspect of machine learning systems [20]. Thus, the process is divided into two major steps: dimensionality reduction, and risk estimation in the low-dimensional space.

## Dimensionality Reduction

We address the visibility of the risk estimation system by dimensionality reduction of the input's 14D space onto a 2D space, using the well-known t-SNE method [10]. The aim is to obtain a clear and visible separation between the high-risk and the low-risk contracts. We divide the data into training and testing data sets, with 20000 and 10000 contracts, respectively. All data manipulations, which include dimensionality reduction and machine learning system training, were applied on the training data set, and afterwards applied and tested on the testing set. Figure 1 presents the $R^{14} \rightarrow R^2$ dimensionality reduction for several values of *perplexity* which is a parameter of the t-SNE method. From this picture, two tendencies are visually observed: (1) as the perplexity increases in value the 2D map becomes more compact, and (2) the separation between high and low risk regions enhances. The second tendency is the one we are looking for. Although, at this stage, the conclusion about the separability of high/low risks is a result of the visual inspection only, it will be confirmed numerically during the next stages.

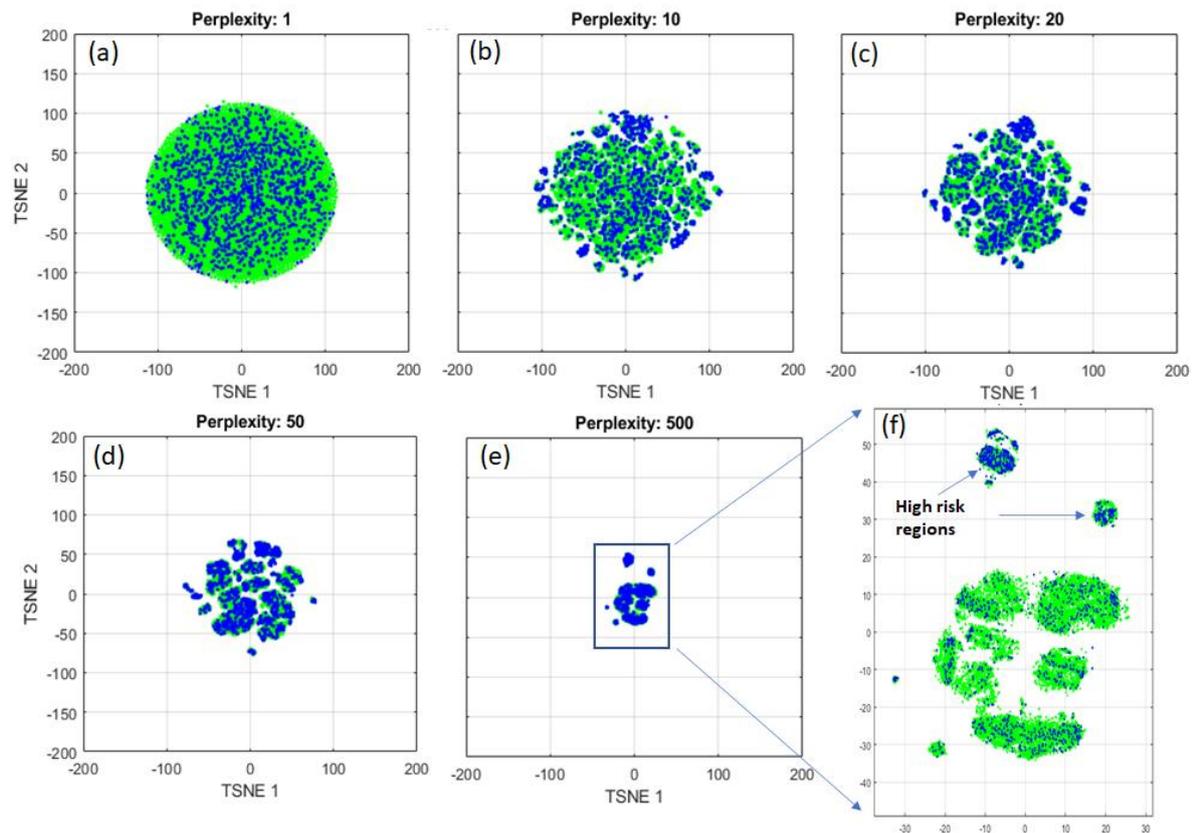

**Figure 1:** Training set $R^{14} \rightarrow R^2$ dimensionality reduction for several perplexity values. (a), (b), (c), (d), (e) – correspond to the perplexity values: 1, 10, 20, 50, 500. (f) – zoom-in of (e). Green points represent contracts with no claims, blue points represent contracts with claims. Regions with a high concentration of blue points correspond to regions with higher risk.

t-SNE is applied on the training set and provides a point-to-point mapping. This form of mapping is not a general transformation which can be applied to a new arbitrary point in the $R^{14}$ space. As a result, the mapping cannot be used to transfer previously unseen test data. We overcome this hurdle by using a standard NN. The NN's structure is as follows: input layer with 14 neurons, a single hidden layer with 100 neurons, and a 2-neuron output layer which corresponds to the two-dimensional embedding of t-SNE. We used a *tanh* transfer function for the hidden layer and a *linear* function for the output layer (See Figure 2). To obtain a general mapping (in contrast to the t-SNE point-to-point mapping presented earlier), a NN was trained on the training set of the t-SNE mapping with perplexity 500. This results in a $R^{14} \rightarrow R^2$ transformation denoted by $NN_{tsne}$.

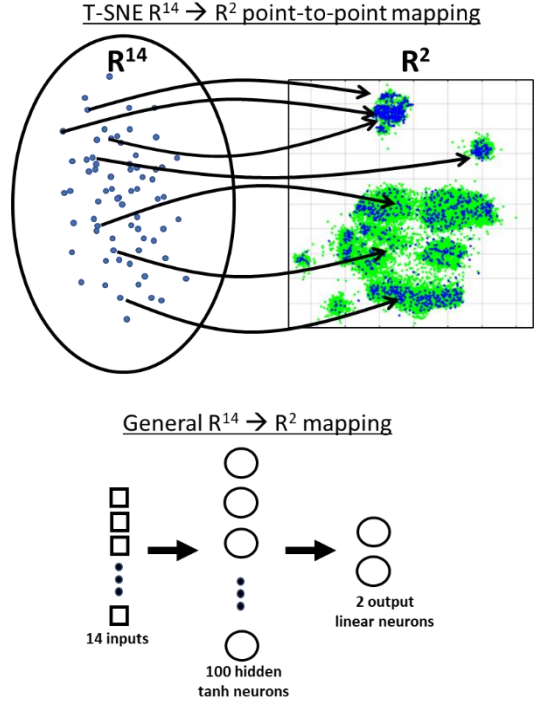

**Figure 2:** t-SNE point-to-point mapping and a general neural network transformation which models it ($NN_{tsne}$).

# 3. Results

**Risk Estimation Function**

The high-risk and the low-risk regions can be visually inspected quite easily from the images in Figure 3 but to estimate the risk numerically, we need a function which associates a contract with its risk value. To estimate this function, we choose a NN (denoted $NN_{risk}$) with a similar structure as in Figure 2 but with 2 inputs, 5 hidden and 1 output neurons. Although our target consists of zeros and ones (an absence, and an appearance of a claim, respectively), because $NN_{risk}$ has a continuous non-limited output, it can have any value inside, as well as outside the [0,1] interval. This output is normalized to a [0,1] interval, which represents a risk estimated by the system (note: in the current realization, the NN original output laid in the [0.0371, 0.2295] interval, which, as mentioned previously, has been normalized to the [0,1]). The result is a continuous function that reproduces well the regions in $R^2$ that contain the data (green and blue points), however, it fails in the intermediate regions that contain no data (Figure 4 (a)). To overcome this drawback, we introduced a discretized version of the risk function, which results from a 100x100 discretization of $R^2$. We take into account only the pixels which contain the underlying points, while zeroing the empty regions (Figure 4 (b)). This results in a surface with several "single-pixel peaks", which are removed by averaging with a 3x3 smoothing kernel (Figure 4 (c)). The final continuous surface is referred to as a risk surface. Note that because the zero-valued intermediate regions do not represent a real zero-risk, but an absence of training data, we must pay a careful attention to the test points which can appear in these regions.

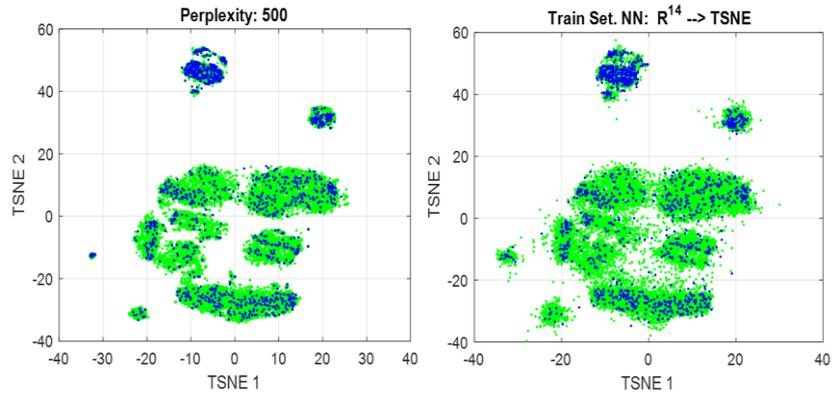

**Figure 3:** t-SNE with perplexity = 500 (left) and $NN_{tsne}$ mapping of the training (right). Green points denote contracts with no claims, blue points denote contracts with claims.

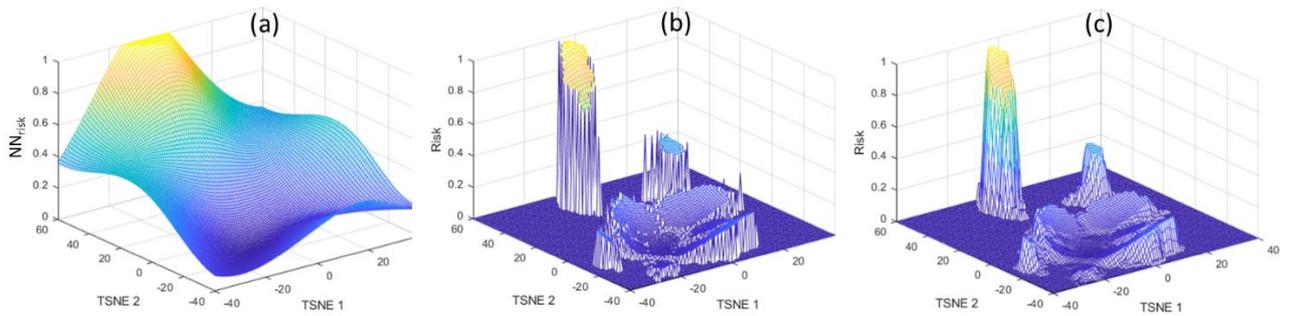

**Figure 4:** Risk surfaces. (a) $NN_{risk}$ $R^2 \to [0,1]$ continuous surface, (b) discretization of $NN_{risk}$ over existing training points, (c) smoothened version of (b).

## Testing

To test our model, we apply the following process on the test data:
1. Normalization of the test data according to the normalization parameters which were used for the training data.
2. Dimensional reduction $R^{14} \to R^2$, which is achieved by applying $NN_{tsne}$
3. Risk calculation using the risk surface.

As mentioned earlier, the test set contains 10000 contracts. After mapping them onto a t-SNE plane, 51 contracts appear to fall out of the risk surface and as a result were removed from consideration, leaving 9949 contracts for testing. The results of this process are shown in Figure 5, where we can see two segments, the medium and high risk regions, significantly separated from the majority (Figure 5 (b), (c)).

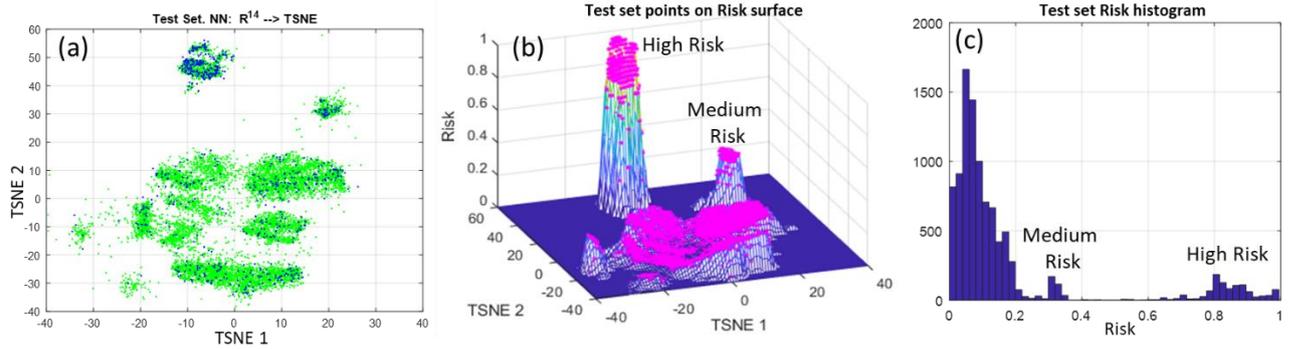

**Figure 5:** Test set result. (a) $NN_{tsne}$ $R^{14} \rightarrow R^2$ mapping into t-SNE plane, (b) t-SNE points on the risk surface, (c) histogram of the testing risk values.

To validate our risk estimation result, we calculated:
1. Risk – claim correlation.
2. Statistics of contracts and claims beyond a continuous thresholding over the risk.

Pearson's correlation of our risk with the real claims is 0.204. Out of the 9949 test contracts 738 have claims, which sets the ratio of claims in the full test dataset to 7.4%. If we divide the risk into three groups, [0,0.3), [0.3,0.5), [0.5,1], associated with low, medium, and high risk (Figure 5(c)), we will get 8592, 302, and 1055 contracts with 464, 38, and 236 claims respectively, which are 5.4%, 12.6%, and 22.4% of the total amount of contracts in these groups. Note that the separation into three risk groups was done according to the estimated risk, which appears to fit this real-life situation. This procedure gives the insurance company several advantages: (a) an improved risk picture of the portfolio, (b) allows for a better pricing strategy, (c) the insurer can remove high-risk contracts from the portfolio.

In Figure 6 we present the result of the continuous thresholding, where policies beyond a specific threshold are grouped together and their amount, together with the ratio of claims are calculated. The results show an almost proportional increase of the claim ratio associated with an increase of the estimated risk. The fluctuations which start beyond the risk value of 0.85, are associated with insufficient statistics because of a small number of contracts, as can be seen in the blue line.

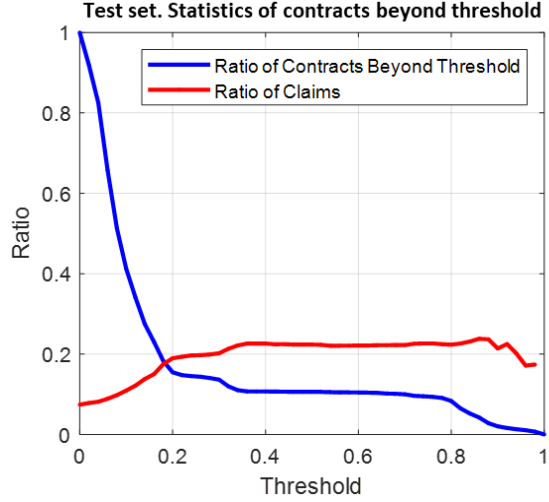

**Figure 6:** Test set. Statistics of contracts beyond continuous thresholding of risk. Blue line denotes the ratio of contracts beyond a specific threshold. Red line denotes the claim ratio in the contracts beyond a specific threshold.

**Using other ML methods**

The use of a neural network to obtain the final risk ($R^2 \to [0,1]$) is not obligatory and essentially any model which is general enough will be sufficient. To this end we present several alternative models which are trained on the reduced two-dimensional space to predict the claim risk. In addition to the neural network used, we considered linear and logistic regression, random forest, and gradient boost. The performance on the test set was measured via the AUC value and is presented in Table 1.

| Model | AUC | Tuning Parameters |
|---|---|---|
| Linear Regression | 0.6298 | None |
| Logistic Regression | 0.6305 | None |
| Neural Network | 0.6703 | Single hidden layer with a *tanh* activation and 10 neurons, and a linear output layer |
| Random Forest Regressor | 0.6592 | 10 trees with max. depth of 5 |
| Gradient Boost Regressor | 0.6676 | 15 trees with max. depth of 5 |

**Table 1:** Risk estimation in t-SNE 2D space ($R^2 \to [0,1]$). Model performance with description of hyperparameters.

Each non-linear model was trained after a grid-search for the optimal choice of hyperparameters. For models with a continues unbounded output value, we normalized the resulting risk to a [0,1] range. As we can see from the table, the non-linear models outperform the linear ones. While the neural network gives the best performance, the other non-linear models are still considerably close to it. Although in general the values of the AUC presented here are quite low, these are very representative of the results in the field of auto insurance [21], [22].

To further establish the robustness of our approach we compare the results in Table 1 to the same set of models without the stage of dimensional reduction, i.e., the models were trained directly on the high dimensional space ($R^{14} \rightarrow [0,1]$). As before, the model parameters were tuned for optimal performance with the results shown in Table 2. From this Table we conclude that although the non-linear models score higher than the linear models, the differences here are significantly smaller than in Table 1. In addition, we can see that even though the $R^{14} \rightarrow [0,1]$ risk estimation is better than the $R^2 \rightarrow [0,1]$, the difference remains small - especially in the NN model case, which presents a very robust behavior and almost no degradation in the performance.

| Model | AUC | Tuning Parameters |
|---|---|---|
| Linear Regression | 0.6701 | None |
| Logistic Regression | 0.6704 | None |
| Neural Network | 0.6777 | Single hidden layer with *tanh* activation and 100 neurons, and a linear output layer |
| Random Forest Regressor | 0.6811 | 20 trees with max. depth of 5 |
| Gradient Boost Regressor | 0.6747 | 20 trees with max. depth of 3 |

**Table 2:** Risk estimation in the original high-dimensional space ($R^{14} \rightarrow [0,1]$). Model performance with description of hyperparameters.

**Comparison to the Insurer's risk**

Since we know the premium and the vehicle value of each contract, we can calculate a **premium / vehicle value** ratio which we are hypothesizing should be proportional to the risk as it was estimated by the original insurer. After normalizing this value to a [0,1] range, we will refer to it as the "Insurer Risk" and denote it by $R_{insurer}$. The risk surface which is based on $R_{insurer}$ is presented in Figure 7 ((a), (b)). (Note: The $R_{insurer}$ surface was produced exactly in the same manner as our risk surface, including zeroing pixels with non-existing data points and the surface smoothing with a 3x3 kernel). For a clear view of the surface, we present it from two distinct points of view. The $R_{insurer}$ risk distribution is given by the histogram in Figure 7(c). Comparison of these results with our risk (Figure 5) shows a different surface structure, and a very different histogram. The $R_{insurer}$ has no clear separation into risk groups, and its histogram is monotonic and long tailed. Indeed, one can argue that maybe the $R_{insurer}$ is more accurate than our risk. To examine this argument, we calculated for $R_{insurer}$ its Pearson's correlation with the real claims and its AUC value, which correspond to 0.110 and 0.601 respectively. These values are lower than ours which, as already stated, are 0.204 and 0.670. In addition, we compare the result of the continuous thresholding of $R_{insurer}$ and our model's risk. Figure 8 presents this continuous thresholding and emphasizes a specific point associated with 20% of the contracts. This group, which represents 20% of the riskiest contracts (according to the calculated risk), contains 16.4% of claims for our risk calculation but only 12.1% of claims for the Insurer's risk. This means that our tool is capable in identifying risky contracts better than the original insurer. Table 3 presents these values for several percentages of contracts and shows that this tendency (our risk estimation is better than the $R_{insurer}$) is consistent throughout.

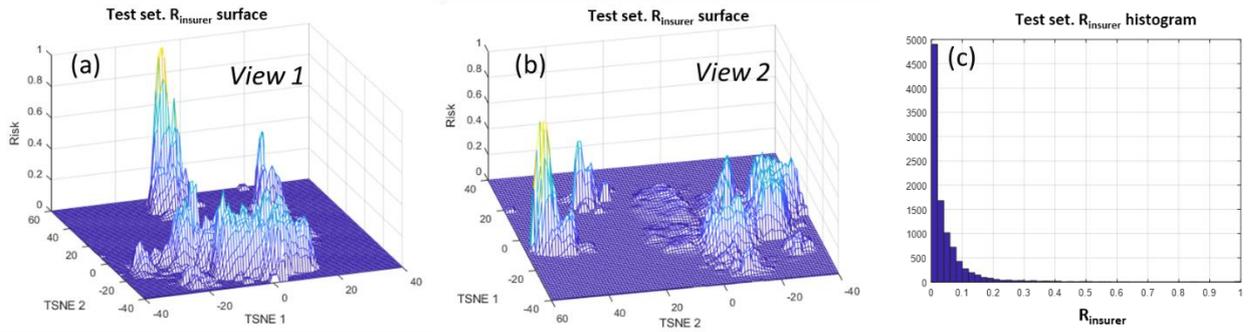

**Figure 7:** Test set. $R_{insurer}$ surface and histogram. (a) and (b) – two points of view of the risk surface, (c) histogram of the insurer's risk distribution.

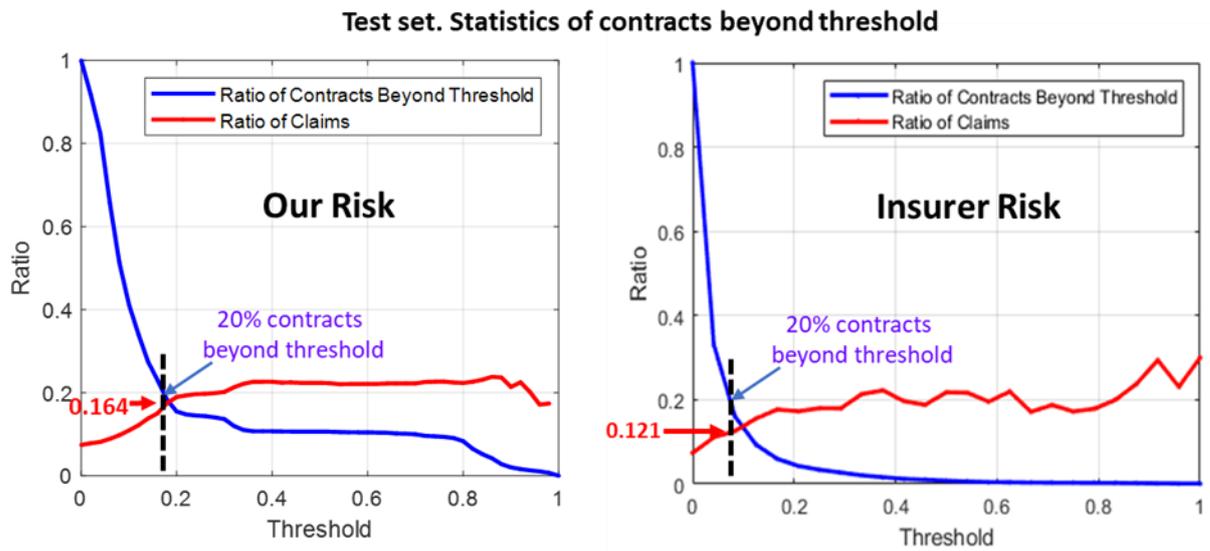

**Figure 8:** Continuous thresholding for our and $R_{insurer}$ risks.

| % of contracts | % of claims (our risk) | % of claims (Insurer's risk) |
|---|---|---|
| 100 | 7.40 | 7.40 |
| 50 | 9.92 | 9.38 |
| 20 | 16.37 | 12.08 |
| 10 | 22.26 | 14.86 |

**Table 3:** Percentage of contracts and claims for our and $R_{insurer}$ risks.

**Visual Assistance and Validation Tool**

Because of the visibility of our method, it can be used as an assistance tool to validate the results derived from other methods. Figure 9 shows the risk surfaces of two methods, together with a choice of three contracts whose risk we designate by three colored points. Note that the risk values of the green points are comparable between the two methods (low value for both), while the red and the magenta points are not. This means that the insurer can validate his results by comparing it to our results and take precaution steps in case of disagreement between the methods. One can argue that there is no need to use our method as a validation tool, and that any other risk estimation method can serve this function. Theoretically this is correct, but in practice this is not feasible for an arbitrary method. The reason behind our argument lies in the regularity and relative smoothness of our risk surface, versus the apparent non-regularity of the $R_{insurer}$. Consequently, the use of $R_{insurer}$ as a validation tool will produce an instability of the results since geometrically close points can have very different risk values, and thus severely compromise its reliability.

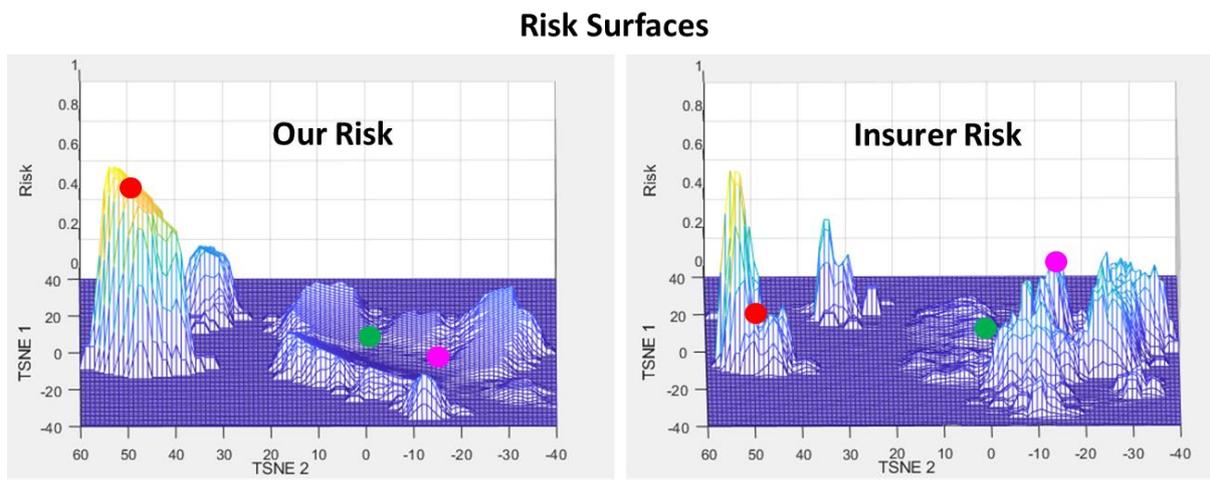

**Figure 9:** Visual Assistance and Validation. Risk surfaces with three different color marked contracts. Green point values calculated by both methods are comparable (low value for both methods), while the risk of the red and the magenta points are not.

# 4. Discussion

In this work, we have presented an auto insurance risk estimation method which takes advantage of t-SNE based visualization together with a neural network. Using data from an auto insurance company we have demonstrated that the complexity of the risk can be reduced to a smooth two-dimensional surface. The surface exhibited clear and visible separation into three distinct risk groups, which allow the insurer a visual validation of the risk estimation. For example, by making a direct estimation from the original $R^{14}$ space, one can now validate whether this estimation is consistent with an appropriate region in $R^2$. We compared the performance of our method to the performance of the original Insurer's risk estimation and showed that indeed our method is superior in separating the risk groups.

We should point out that the notion of visibility, a predominant part in the structure of our method, is different from the notion of explainability, which aims to explain the influence of specific parameters on the risk estimation. We intend to address this issue in future research. We also intend to investigate other dimensional reduction techniques such as UMAP [23], as well as the risk surface construction method, which might have an influence on the final results.